# V2V-based Collision-avoidance Decision Strategy for Autonomous Vehicles Interacting with Fully Occluded Pedestrians at Midblock on Multilane Roadways

**Fengjiao Zou, SM. ASCE[1*]; Hsien-Wen Deng[2]; Tsing-Un Iunn[3];
Jennifer Harper Ogle, Ph.D., M. ASCE [4]; and Weimin Jin, Ph.D.[5]**

[1]Department of Civil Engineering, Clemson University, 110 Lowry Hall, Clemson, South Carolina, 29634, USA, Email: fengjiz@clemson.edu
[2]Nan Ya Plastics Co. America, South Carolina, 29560, USA, Email: stevenxdeng@gmail.com
[3]TSMC, Phoenix, Arizona, 85083, USA, Email: tiunn@clemson.edu
[4]Department of Civil Engineering, Clemson University, 29634, USA, Email: ogle@clemson.edu
[5]Arcadis Inc. Houston, 77494, USA, Email: weiminj@clemson.edu

**ABSTRACT**

Pedestrian occlusion is challenging for autonomous vehicles (AVs) at midblock locations on multilane roadways because an AV cannot detect crossing pedestrians that are fully occluded by downstream vehicles in adjacent lanes. This paper tests the capability of vehicle-to-vehicle (V2V) communication between an AV and its downstream vehicles to share midblock pedestrian crossings information. The researchers developed a V2V-based collision-avoidance decision strategy and compared it to a base scenario (i.e., decision strategy without the utilization of V2V). Simulation results showed that for the base scenario, the near-zero time-to-collision (TTC) indicated no time for the AV to take appropriate action and resulted in dramatic braking followed by collisions. But the V2V-based collision-avoidance decision strategy allowed for a proportional braking approach to increase the TTC allowing the pedestrian to cross safely. To conclude, the V2V-based collision-avoidance decision strategy has higher safety benefits for an AV interacting with fully occluded pedestrians at midblock locations on multilane roadways.

**Key Words:** Autonomous vehicle (AV); Fully occluded pedestrian; Collision-avoidance decisions; Time-to-collision (TTC); Vehicle-to-vehicle (V2V) communication

## INTRODUCTION

One of the safety challenges for autonomous vehicles (AVs) in the absence of connectivity
is occluded pedestrians because AVs could not detect the occluded pedestrians in time to take evasive actions (Shetty et al., 2021). Figure 1 depicts one hazardous occlusion situation of a pedestrian crossing midblock on a multilane road at an unmarked location. As vehicle 1 in the rightmost lane slows down for the pedestrian, it also blocks the pedestrian. Vehicle 2 in the adjacent lane may not detect the pedestrian due to the occluded area and the lack of pedestrian crossing markings, nor can the pedestrian see the vehicle. In such a situation, vehicle 2 may be unable to stop or slow down to avoid striking the pedestrian. Thus, pedestrian crossing maneuvers at unmarked midblock locations on multilane roads present a challenging driving context for AVs. However, prior research indicates that supplementing AVs with vehicle-to-vehicle (V2V) technology could address scenarios such as obscured pedestrians crossing at non-junction locations (Katrakazas et al., 2015). An AV that receives information through V2V can aggregate these messages and make appropriate decisions, like avoiding possible collisions with pedestrians, even for scenarios in which an AV's sensors cannot detect pedestrians due to an occlusion. Through V2V communication, an AV can obtain the actual state of obstacles by connecting with surrounding vehicles to provide accurate long-term risk assessment (Katrakazas et al., 2015).

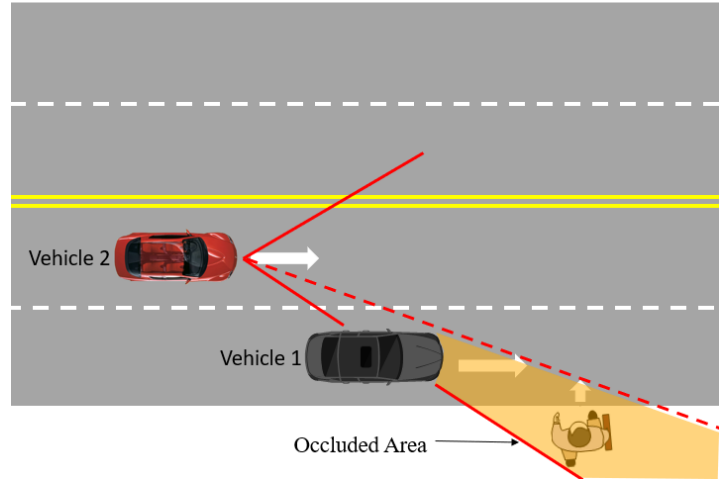

**Figure 1. Scenario with pedestrian fully occluded**

Detecting fully occluded pedestrians earlier and faster is crucial for AV collision-avoidance decisions because it affects how much time an AV has to brake to avoid a potential collision with an occluded pedestrian. The previous studies regarding AVs and occluded pedestrians were conducted primarily on 2-lane roads rather than more complicated unmarked midblock locations on multilane roads. Besides, AV response in most studies considered full braking in the collision-avoidance decision strategy, causing a subject AV to brake dramatically and, thus, increasing the collision risk between an AV and its immediate upstream vehicles in the same lane. This paper contributes to previous studies by considering the following: 1) utilization of V2V communication technology to address the problem of pedestrians being fully occluded by moving vehicles at midblock on multilane roads, 2) development of a V2V-based collision-avoidance decision strategy for an AV to improve its safety, and 3) the introduction of a proportional braking response to allow for an appropriate yet less drastic deceleration response to the pedestrian when warranted.

**LITERATURE REVIEW**

Pedestrian occlusion (partially or fully) is common in urban areas. They can be occluded by infrastructures, parked vehicles, or moving vehicles (El Hamdani et al., 2020; Shetty et al., 2021). When a pedestrian is occluded, it becomes a challenge for an AV's mounted sensors (Gilroy et al., 2021). Some researchers fused camera and radar sensors to address the challenge of pedestrians darting out behind parked vehicles (Palffy et al., 2019). Others fused lidar and radar sensors for partially occluded pedestrian detection (Kwon et al., 2017). Overall, sensor fusion schemes had better detection performance than sensors alone. However, they can only deal with partially occluded pedestrians and cannot detect fully occluded ones (El Hamdani et al., 2020).

Vehicular communication brings a new dawn to the safety of occluded pedestrians. Vehicle-to-infrastructure (V2I) communication enables vehicles to communicate with roads equipped with roadside units (RSU). Some researchers used cameras and laser scanners as RSU to detect and send pedestrian sidewalk information to vehicles through V2I (Goldhammer et al., 2012). Using sensors (e.g., radar, camera) installed at intersections, V2I provides a bird's eye view of the current scene and can recognize the occluded pedestrian earlier. However, most V2I applications are implemented at intersections. As most (73%) pedestrian crashes happened at midblock locations (NHTSA, 2018), V2I communication technology has limited safety effects on midblock locations without crosswalks. Another vehicular communication application is vehicle-to-pedestrian (V2P). Pedestrians can carry mobile phone devices and send signals to surrounding vehicles. Vehicles receive an alert if a collision is probable, even when the pedestrian is occluded (Wu et al., 2014). However, there are many issues concerning V2P safety applications, such as enabling many smartphones with communication capability. Also, the most vulnerable pedestrians (i.e., children, seniors, and those with disabilities), may not have or be able to afford smartphones.

V2V communication has become more practical for occluded pedestrian detection and is being tested as a potential solution due to the rapid development of wireless communication technology and computing speed. Researchers have been worked on fusing V2V with existing vehicle advanced driver assistance systems (ADAS), like pre-collision systems (PCS) (Liu et al., 2015) and pedestrian automatic emergency brake systems (PAEB) (Flores et al., 2019; Li et al., 2017; Tang, 2015) to enable information sharing and enhance pedestrian safety. García et al. fused laser scanners, camera sensors, and V2V communication to provide accurate pedestrian tracking, thus avoiding pedestrian occlusion (García et al., 2013). Similarly, Flores et al. fused lidar sensing with V2V to develop a pedestrian collision avoidance system (Flores et al., 2019). Tang et al. combined V2V and PAEB to protect occluded pedestrians (Tang et al., 2016). They tested the proposed V2V-PAEB architecture in the PreScan simulation environment and used a scenario of a pedestrian occluded by a truck when crossing at an intersection. Results showed that the proposed V2V-PAEB system detected the pedestrian earlier than the case without V2V communication and avoided collision with the pedestrian. Buy Tang et al. used one low vehicle speed (15 mph) in the simulation and did not investigate the effects of the vehicle speed on the V2V-PAEB system.

To summarize the literature listed above, decision strategies utilizing vehicular communication, especially V2V communication, can improve occluded pedestrian safety compared to decision strategies utilizing in-vehicle sensors only. Most studies considered fused V2V with full braking in the ADAS systems, causing a subject AV to brake dramatically and potentially increasing the collision risk between an AV and its immediate upstream vehicles in the same lane. Besides, most studies considered pedestrians partially occluded by parked vehicles or static infrastructures. This paper considered a scenario in which a pedestrian is fully occluded by an AV's downstream vehicles in adjacent lanes. Further, midblock locations were less considered in those research than in intersection scenarios. Also, few researchers considered the impacts of different vehicle speeds on the system's performance. This paper addresses several research gaps in the prior literature by proposing a V2V-based collision-avoidance decision strategy for AVs to improve fully occluded pedestrian safety; introducing a proportional braking response to allow for an appropriate yet less drastic deceleration response to the pedestrian when warranted; testing system performance across an array of vehicle approach speeds (10 mph to 70 mph); and using a midblock crossing location with pedestrian fully occluded by moving vehicles in adjacent lanes.

**METHODS**

Figure 2 shows the framework of the V2V-based collision-avoidance decision strategy. The top portion represents the functions in the transmitter vehicle (vehicle 1), and the bottom part displays the functions in the subject AV (vehicle 2). In this study, both the subject AV and the transmitter vehicle are equipped with V2V communication modules. Taking Figure 1 as an example, the transmitter vehicle (vehicle 1) can detect the pedestrian, while the subject AV (vehicle 2) does not detect the occluded pedestrian. In Figure 2, V2V communication technology builds the bridge between the transmitter vehicle and the subject AV to allow communication. The transmitter vehicle utilizes sensors to detect a crossing pedestrian. Once the transmitter vehicle detects a pedestrian, it broadcasts its position and velocity to the subject AV through V2V communication. The subject AV is responsible for receiving the pedestrian position and velocity from the transmitter vehicle and decision-making while interacting with possible pedestrian collision scenarios during occlusion. In Figure 2, each time the subject AV obtains the pedestrian position and velocity from the transmitter vehicle, it computes a time-to-collision (TTC) between itself and the occluded pedestrian. Suppose the computed TTC is less than a predefined TTC threshold; the subject AV will need to reduce speed to avoid hitting the pedestrian. In that case, the subject AV will undertake braking action proportional to collision potential as advised by the adaptive collision-avoidance system, thus, mitigating a collision with the pedestrian.

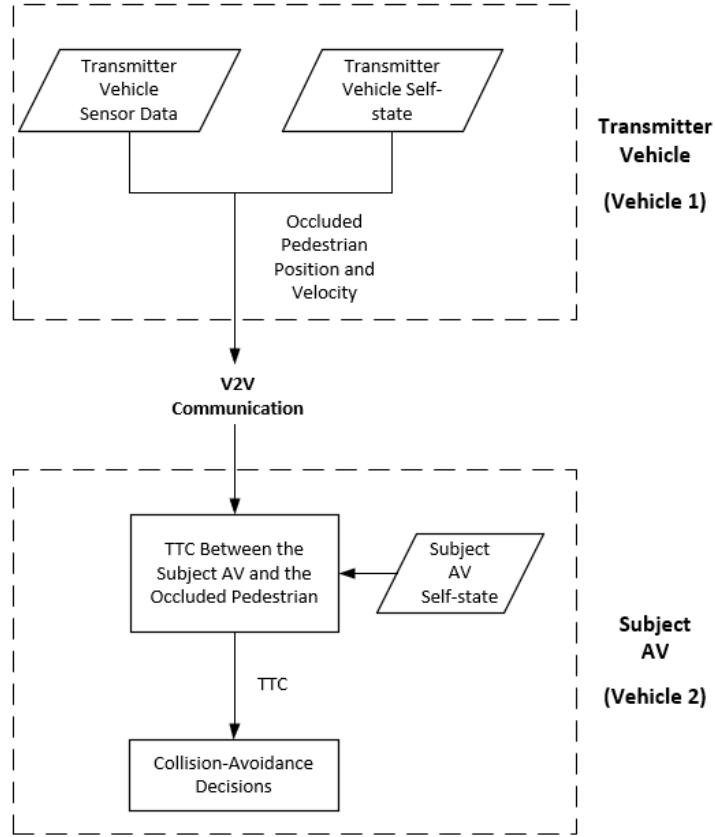

**Figure 2. V2V-based Collision-Avoidance Decision Strategy**

*TTC Between the subject AV and the occluded pedestrian*

In the motion planning module of an AV, risk assessment is crucial to mitigate collisions between an AV and its surrounding environment, including vehicles and pedestrians. Risk assessment can be performed using risk indicators, such as the TTC (J. R. Ward et al., 2015). Typically, the TTC refers to car-following situations only, where TTC is calculated between a lead and the following vehicle. This paper calculates the TTC between the subject AV and the pedestrian in a manner similar to previous studies (Karamouzas et al., 2009; Steven M. LaValle, 2006). The researchers assume that both the subject AV and the pedestrian are spheres with positions $(\overrightarrow{X_V}, \overrightarrow{X_p})$, velocities $(\overrightarrow{v_V}, \overrightarrow{v_p})$, and radius. The definition of the vehicle's "sphere" radius ($R_v$) is half of the vehicle's body length, and the radius of the pedestrian "sphere" ($R_p$) is the horizontal distance from the center of the pedestrian's head to the end of the most extended finger when the arms are open. Table 1 lists all the notations used in the method section.

**Table 1** Model Notation

| Symbol | Meaning |
| --- | --- |
| $\vec{X_p}$ | Pedestrian position |
| $\vec{X_V}$ | AV position |
| $\vec{X}$ | Relative position between AV and pedestrian |
| $\vec{v_p}$ | Pedestrian velocity with direction |
| $\vec{v_V}$ | AV velocity with direction |
| $\vec{v}$ | Relative velocity between the AV and the pedestrian |
| $R_p$ | Radius of the pedestrian (treated as a sphere) |
| $R_V$ | Radius of the AV (treated as a sphere) |
| $R_r$ | Relative radius between the AV and the pedestrian |
| $\tau$ | Time-to-Collision (TTC) |
| $\tau_{max}$ | Threshold of TTC for decision |
| $a$ | Coefficient of a quadradic equation $ax^2 + bx + c = 0$ |
| $b$ | Coefficient of a quadradic equation $ax^2 + bx + c = 0$ |
| $c$ | Constant of a quadradic equation $ax^2 + bx + c = 0$ |
| $D$ | Discriminant of a quadradic equation $ax^2 + bx + c = 0$ |
| $P_B$ | Breaking pressure |
| $P_{max}$ | Maximum braking pressure |
| BFFS | Basic Free Flow Speed on rural, 75 mph |
| $a_b$ | Maximum Comfortable deceleration, 11.2 $ft/s^2$ |

The objective is to determine the minimum possible TTC ($\tau$), which can be compared to a predefined threshold value ($\tau_{max}$) at which there is no potential for collision. Figure 3 shows an impending collision with $\tau$ approaching zero as the subject AV and pedestrian spheres collide. To find the $\tau$, the relative positions, relative velocities, and relative radius of the pedestrian and the subject AV are defined below:

$$\vec{X} = \vec{X_p} - \vec{X_V} \tag{1}$$
$$\vec{v} = \vec{v_p} - \vec{v_V} \tag{2}$$
$$R_r = R_p + R_V \tag{3}$$

The subject AV and the pedestrian would collide in time $\tau$ when Equation 4 is met:

$$|\vec{X} + \vec{v}\tau| = R_r \tag{4}$$

To solve this equation, expand Equation 4 by the quadratic formula below:

$$(\vec{v} \cdot \vec{v})\tau^2 + 2(\vec{X} \cdot \vec{v})\tau + \vec{X} \cdot \vec{X} - R_r^2 = 0 \tag{5}$$

Then, the coefficients and discriminant of $\tau$ in Equation 5 can be determined below:

$$a = \vec{v} \cdot \vec{v} \tag{6}$$
$$b = \vec{X} \cdot \vec{v} \tag{7}$$
$$c = \vec{X} \cdot \vec{X} - R_r^2 \tag{8}$$
$$D = b^2 - ac \tag{9}$$

Thus, the calculation of $\tau$ is derived by solving Equation 5 as follows:

$$\tau^{\pm} = \frac{-b \pm \sqrt{D}}{a} \tag{10}$$

However, not every scenario generated a valid value of $\tau$. For instance, when no pedestrian is detected, indicating no collision will occur; thus, there will be no valid TTC ($\tau$). This paper sets 10000 seconds as a number representing infinity TTC ($\tau$), indicating no valid TTC. Sometimes there are two solutions $(\tau^-, \tau^+)$ after solving Equation 5. In that case, then the smaller valid TTC $(\tau^-)$ will be used for the subject AV. Equations 11 to 15 determine the TTC used for the decision strategy:

$$D < 0, No\ valid\ TTC \tag{11}$$
$$\tau^\pm < 0, No\ valid\ TTC \tag{12}$$
$$\tau^- = 0\ or\ \tau^+ = 0, TTC = 0 \tag{13}$$
$$\tau^- < 0, \tau^+ > 0, TTC = \tau^+ \tag{14}$$
$$\tau^\pm > 0, TTC = \tau^- \tag{15}$$

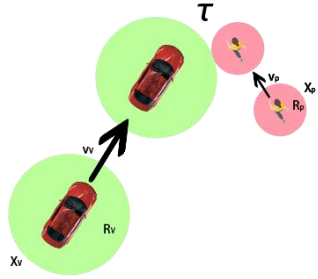

**Figure 3. Assumption of collision in time $\tau$**

*Collision-Avoidance Decisions*

To reduce the potential collision risk between a subject AV brake reaction and any immediately following vehicle in the same lane, the collision-avoidance decision strategy proactively detects a potential collision and reduces the need for dramatic braking. Suppose a collision is presumed to occur in time τ. In that case, the subject AV can generate a braking force (F) to neutralize the original velocity (Karamouzas et al., 2009), as shown in Figure 4, then position. Therefore, at the same time $\tau$, a collision would be avoided. In the simulation, there is no direct measurement of the braking force; rather, the braking force is accomplished via throttle ratio and braking pressure. Large TTC values indicate the subject AV is not close to a collision, and a slight braking action would avoid a potential collision. However, a small TTC would indicate the subject AV is close to a collision and the subject AV needs to exert more effort to avoid a potential collision. Under random traffic conditions, there would presumably be scenarios in which a physical reaction to the V2V alerts would not be possible, thus resulting in a collision. This paper tested only scenarios with the potential for collision avoidance.

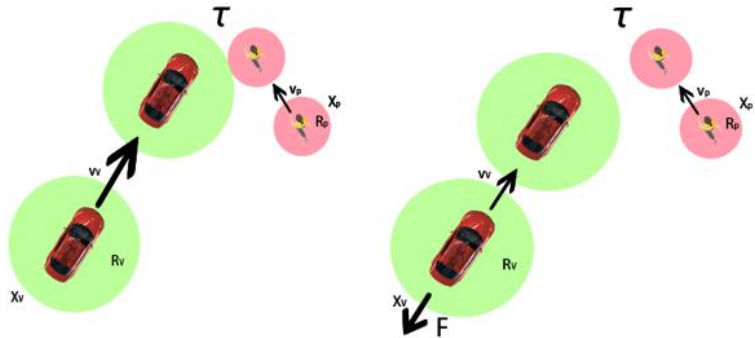

**Figure 4. Collision avoided at the same time $\tau$ by applying braking force F**

The subject AV's decisions depend on the following two constraints: $\tau_{max}, P_{max}$, where $\tau_{max}$ is a predefined threshold set to allow avoidance responses at speeds up to 75 mph, and $P_{max}$ is the maximum braking pressure. A $\tau_{max}$ of 10 seconds (Equation 16) allows the subject AV to stop at a comfortable deacceleration (Greibe, 2007) while traveling at a rural base free-flow speed of 75 mph (TRB, 2010). Therefore, the researchers apply the threshold of 10 s for $\tau_{max}$.

$$\tau_{max} = \frac{BFFS}{a_b} = 10\ s \tag{16}$$

There will be a decision only if TTC is not larger than $\tau_{max}$, then a new braking pressure for actuation is generated following Equation 18. Otherwise, the subject AV does not need to react and will maintain the current speed (Equation 17). For example, assuming a subject AV has a full brake $P_{max}$ of 200 bars (2900 psi) (SAE International, 2015), this AV's maximum braking pressure is 200 bars when the vehicle applies the full brake. If at timestep zero, the computed TTC is six seconds, then a braking pressure $P_B$ of 40% of the full brake $P_{max}$ or 80 bars would be applied to avoid collision with the occluded pedestrian. After applying 80 bars of braking pressure, the subject AV's speed will be reduced at the next timestep, thus resulting in a higher TTC and a smaller braking pressure. TTC, braking pressure, and speed of the subject AV are computed at each timestep to assist in decision-making.

$$\tau > \tau_{max}, No\ Brake\ Decision \tag{17}$$

$$\tau \leq \tau_{max}, Brake\ by\ applying, P_B = \frac{\tau_{max} - \tau}{\tau_{max}} P_{max} \tag{18}$$

## CASE STUDY

This paper used PreScan, a physics-based simulation platform for AVs (Tideman & Van Noort, 2013), to test the V2V-based collision-avoidance decision strategy for an AV interacting with occluded pedestrians at midblock locations on multilane roads. PreScan can be used to develop AV systems based on sensor technologies like camera, radar, lidar, GPS, and antennas for V2V communication. Figure 5 shows the simulation setup for this research. The car models used in the simulation were a red Mazda RX8 (i.e., subject AV) and a black Audi A8 (i.e., transmitter vehicle). The pedestrian model had a walking speed of 4 ft/s. This paper used 7.3 ft, half of the vehicle's body length, as the Mazda RX8 "sphere" radius and 5 ft as the pedestrian "sphere" radius. This paper used the technology independence sensor (TIS), a sensor group combining the function of a camera and radar in the PreScan simulation. The TIS sensor could detect objects and determine whether it is a pedestrian. Once the pedestrian is detected, the TIS sensor tracks the pedestrian's position and velocity. Researchers used the component of V2X for V2V communication between the transmitter vehicle and the subject AV. The roadway was a four-lane highway with a lane width of 12 feet. No pedestrian crosswalks or refuge islands were available on the road. This paper simplified the simulation by including only the braking model in the AV's controller. So, there was no lane changing or passing maneuvers allowed around the pedestrian in the simulation. All sensors were also assumed to be working correctly.

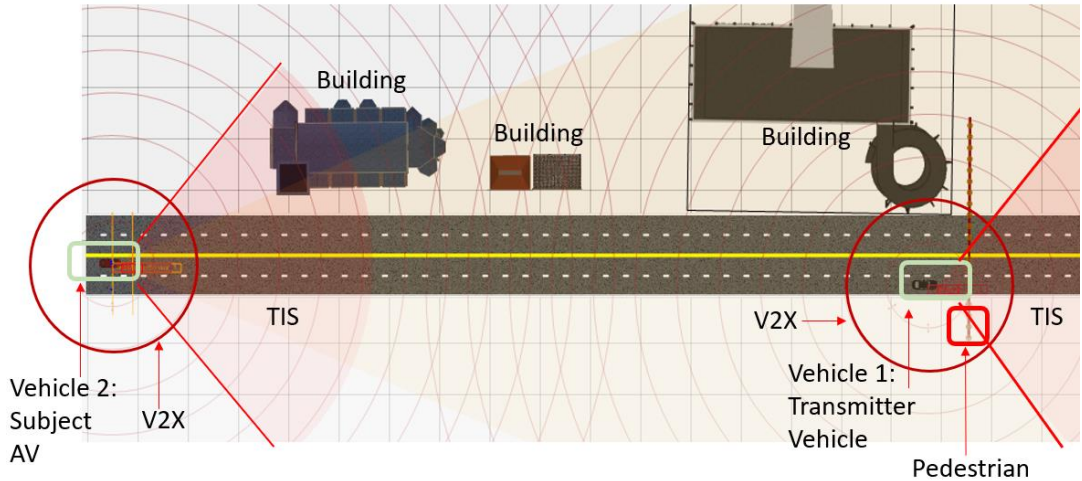

**Figure 5. Simulation setup**

In the simulation, there were two vehicles and one pedestrian on the highway. Figure 6 shows the typical scenario – as the pedestrian begins crossing the road, the subject AV (i.e., red vehicle in the inside/left lane in Figure 6) moves toward the transmitter vehicle (i.e., black vehicle in the outside/right lane in Figure 6) and the pedestrian. Figure 6 portrays the transmitter vehicle detecting the pedestrian crossing the road and occluding the subject AV's view of the pedestrian. The researchers evaluated two decision strategies in this paper: 1) decision strategy with V2V (i.e., the V2V based collision-avoidance decision strategy developed in this paper); 2) decision strategy without V2V (i.e., the base or control scenario considered in this paper). The researchers performed numerous simulations for different subject vehicle speeds varying from 10 mph to 70 mph. The pedestrian crossed using the same trajectory from right to left (outside to inside). In all scenarios, the pedestrian was fully occluded by the transmitter vehicle. Thus, a collision between the subject AV and the pedestrian would occur if subject AV did not have a speed change. The pedestrian entry into the first lane was adjusted based on the subject AV's speed to generate this scenario.

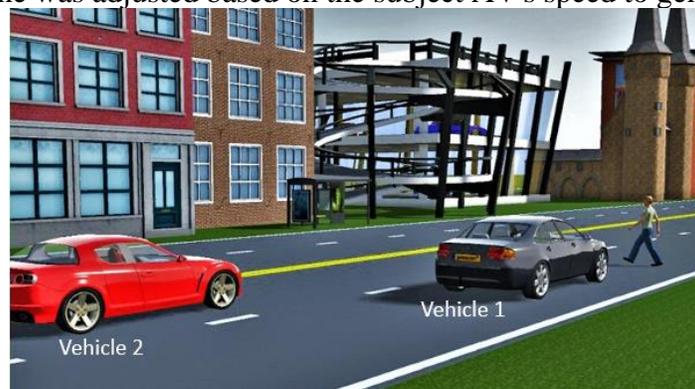

**Figure 6. PreScan simulation scene**

## RESULTS

Table 2 summarizes the outputs of the simulations. The first column in Table 2 contains the speed of the subject AV that varies from 10 mph to 70 mph. The 'detected time' column records the result of the first instance in which the subject AV receives pedestrian information and calculates a TTC. Columns 2-4 provide the outputs of the experimental decision strategy with V2V, while columns 5-7 give the outputs of the control decision strategy without V2V (i.e., the base scenario). The TTC was calculated every 0.02 seconds after the initial V2V detection message receipt or the initial pedestrian detection for the base scenario.

**Table 2** Simulation outputs

| Speed (MPH) | Decision strategy with V2V | | | Decision strategy without V2V | | |
|---|---|---|---|---|---|---|
| | Detected time | TTC | Collision | Detected time | TTC | Collision |
| 1 | 25.88 | 11.9129 | NO | 37.85 | 0.1687 | NO |
| 15 | 23.78 | 12.0777 | NO | 35.42 | 0.1606 | YES |
| 20 | 12.88 | 12.8470 | NO | 24.83 | 0.1117 | YES |
| 25 | 11.63 | 12.6055 | NO | 22.32 | 0.0758 | YES |
| 30 | 10.41 | 10.8220 | NO | 19.37 | 0.0603 | YES |
| 35 | 9.52 | 9.3742 | NO | 17.25 | 0.0677 | YES |
| 40 | 8.63 | 8.2303 | NO | 15.40 | 0.0574 | YES |
| 45 | 7.82 | 7.1975 | NO | 13.78 | 0.0502 | YES |
| 50 | 7.15 | 6.5221 | NO | 12.57 | 0.0383 | YES |
| 55 | 6.50 | 5.9466 | NO | 11.45 | 0.0268 | YES |
| 60 | 5.95 | 5.4241 | NO | 10.47 | 0.0410 | YES |
| 65 | 5.32 | 4.8486 | NO | 9.35 | 0.0108 | YES |
| 70 | 5.02 | 4.5317 | NO | 8.8 | 0.0093 | YES |

The TTC in Table 2 represents the possible collision TTC collected at the first time. In all cases, the experimental decision strategy with V2V produced much larger TTC values than the base scenario, providing sufficient time to undertake braking for collision avoidance. Thus, the decision strategy with V2V outperformed the decision strategy without V2V at all speeds. For all speeds, the average TTC for decision strategy with V2V is 8.642 seconds, while the same average for the base scenario is only 0.0676 seconds. The 'collision' columns in Table 2 indicate whether a collision between the subject AV and the occluded pedestrian occurred in each simulation. Table 2 shows that for the base scenario, collisions occurred in all but one case. Only at a vehicle speed of 10 mph could the subject AV brake successfully to avoid colliding with the pedestrian without V2V communication. However, the vehicle still braked abruptly and stopped right in front of the pedestrian during the simulation. In all cases for decision strategy with V2V, collisions with the pedestrian were avoided.

Figure 7 shows one instance of the differences of the TTC-detected time when the vehicle was proceeding at 45mph for decision strategies with and without V2V. The solid blue line in the upper part of Figure 7 represents the decision strategies with V2V, while the orange dashed line is the base scenario. For both strategies, the initial TTC was set at 10000 seconds, which was a number that researchers set in the algorithm to represent infinity TTC (no valid TTC), indicating no pedestrian was detected or no pedestrian information was received from V2V. The scenario with V2V communication first detected the pedestrian at 7.82 seconds, computed a TTC value of 7.197, and initiated collision avoidance braking to mitigate the occluded pedestrian collision. However, without V2V communication, it took the vehicle 13.78 seconds to detect the pedestrian, which was 5.96 seconds slower. The detected TTC value was only 0.050 seconds, leaving no time for braking actuation; thus, the collision was not avoided.

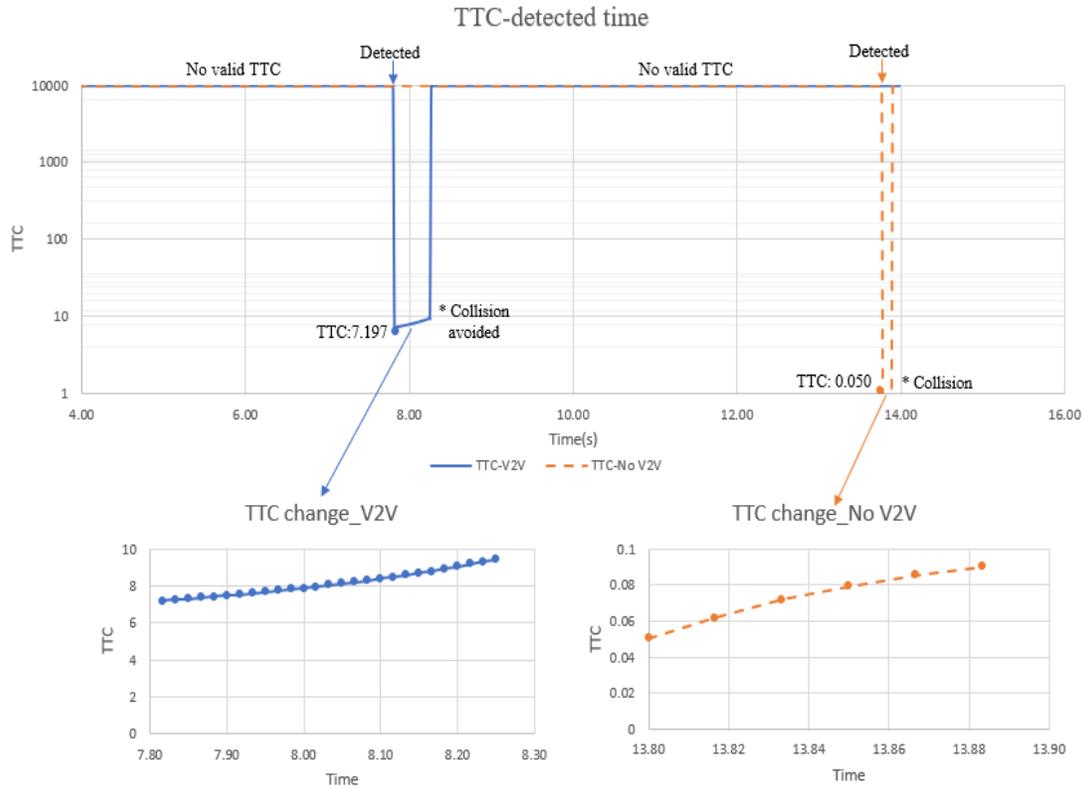

**Figure 7. TTC-detected time for with and without V2V communication (Speed 45mph)**

The bottom two graphs in Figure 7 chart TTC changes after the initial pedestrian detection for both strategies. The solid blue line on the left represents the change in TTC for the decision strategy with V2V communication. It is observed that the TTC increased after the first instance that the subject AV received notice of the pedestrian and computed a TTC at 7.82 seconds. The decision strategy with V2V communication allows the subject AV to apply its brake pedal at each time step if the TTC is less than the specific predefined threshold (i.e., 10 seconds). So, there was less probability of a collision after each successive braking decision, and the TTC increased at each collision avoidance decision until the pedestrian could safely cross. At this time, the TTC returns to the initial infinity TTC (i.e., 10000 seconds), and awaits alerts of the next pedestrian threat. A similar trend of TTC increase is also found in the bottom right graph of Figure 7, which represents the base scenario. However, note the variation in the TTC axis scales – all TTC for the base scenario were minimal (less than 0.1 seconds), leading to collisions between the subject AV and the pedestrian. It was also found that the subject AV began dramatic full-braking for the decision strategy without V2V communication as soon as it sensed the pedestrian. Without V2V communication, even while using max braking pressure (200 bars in this case), the subject AV still collided with the pedestrian because it detected the pedestrian at the very last second. However, with V2V communication, the subject AV started to brake slightly (5.96 seconds) earlier because it received messages from the transmitter vehicle and decelerated in advance. The subject AV with V2V communication never used max deceleration but successfully avoided collisions. The decision strategy with V2V can potentially reduce the collision risk between the subject AV and its immediate upstream vehicles due to decreased intensity and less abrupt braking.

## CONCLUSIONS

To conclude, the increasing trend of midblock pedestrian crashes and the limited capability of AV in-vehicle sensors to detect occluded pedestrians make the pedestrian occlusion issue an urgent problem to solve. This paper investigates the efficacy of AV collision avoidance decision strategies using V2V communication for occluded pedestrian detection at unmarked midblock locations on multilane highways. V2V communication between an AV and its downstream vehicles allows sharing of information regarding midblock pedestrian crossings, thus enabling enhanced collision avoidance. The time-to-collision (TTC) between an AV and a fully occluded pedestrian was evaluated to indicate the collision risk. Based on the collision risk, the researchers developed a V2V-based collision-avoidance decision strategy to mitigate collisions between an AV and an occluded pedestrian for circumstances where the AV can feasibly respond. The decision strategies with and without V2V communication were evaluated in PreScan for different AV speeds varying from 10 mph to 70 mph.

Results show that the V2V-based collision-avoidance decision strategy was successfully tested at all speeds. However, the base scenario resulted in collisions between the AV and occluded pedestrian at all speeds in excess of 10 mph. In all cases, the experimental decision strategy with V2V produced much larger TTC values than the base scenario, providing sufficient time to undertake braking for collision avoidance. Thus, the decision strategy with V2V outperformed the decision strategy without V2V at all speeds. The near-zero TTC for the base scenario indicated no time for the AV to take appropriate action, resulting in dramatic braking followed by collisions. Alternatively, the V2V-based collision-avoidance decision strategy allowed for a proportional braking approach to slow the vehicle and increase the TTC allowing the pedestrian to cross safely. Therefore, the decision strategy with V2V can potentially reduce the risk of secondary collisions with trailing vehicles. This paper reveals that pedestrian occlusion is a challenge for AVs if they rely solely on in-vehicle sensors because all the occluded pedestrian TTCs were near zero without V2V technology. The V2V-based collision-avoidance decision strategy has higher safety benefits for an AV interacting with fully occluded pedestrians at midblock locations on the multilane roadways than the base scenario.

## ACKNOWLEDGMENTS

The authors would like to thank Dr. Mashrur "Ronnie" Chowdhury and Dr. Md Mhafuzul Islam for offering excellent support for this research.